\documentclass[10pt,twocolumn,letterpaper]{article}

\usepackage{cvpr}
\usepackage{times}
\usepackage{epsfig}
\usepackage{graphicx}
\usepackage{amsmath}
\usepackage{amssymb}

\usepackage{algorithm}
\usepackage{algorithmic}
\newtheorem{theorem}{Theorem}
\usepackage{multirow}

\usepackage[pagebackref=true,breaklinks=true,letterpaper=true,colorlinks,bookmarks=false]{hyperref}

\usepackage{color}

\DeclareMathOperator*{\argmin}{arg\,min}

 \cvprfinalcopy 

\begin{document}

\title{Unsupervised Domain Adaptive Object Detection using Forward-Backward Cyclic Adaptation}

\author{Siqi Yang, Lin Wu, Arnold Wiliem, Brian C. Lovell\\
The University of Queensland\\
{\tt\small siqi.yang@uq.net.au, Xiaoxian.wu9188@gmail.com, a.wiliem@uq.edu.au, lovell@itee.uq.edu.au}
}



\maketitle
\thispagestyle{empty}

\begin{abstract}
We present a novel approach to perform the unsupervised domain adaptation for object detection through forward-backward cyclic (FBC) training. 
Recent adversarial training based domain adaptation methods have shown their effectiveness on minimizing domain discrepancy via marginal feature distributions alignment. 
However, aligning the marginal feature distributions does not guarantee the alignment of class conditional distributions.
This limitation is more evident when adapting object detectors as the domain discrepancy is larger compared to the image classification task, \eg various number of objects exist in one image and the majority of content in an image is background. 
This motivates us to learn domain invariance for category level semantics via gradient alignment.
Intuitively, if the gradients of two domains point in similar directions, then the learning of one domain can improve that of another domain.
To achieve gradient alignment, we propose Forward-Backward Cyclic Adaptation, which iteratively computes adaptation from source to target via backward hopping and from target to source via forward passing. 
In addition, we align low level features for adapting holistic color/texture via adversarial training.
However, the detector performs well on both domains is not ideal for target domain. As such, 
in each cycle, domain diversity is enforced by maximum entropy regularization on the source domain to penalize confident source-specific learning and minimum entropy regularization on target domain to intrigue target-specific learning. 
Theoretical analysis on the training process is provided, and extensive experiments on challenging cross-domain object detection datasets have shown the superiority of our approach over the state-of-the-art.

\end{abstract}


\section{Introduction}

\begin{figure}[t]
\begin{center}
   \includegraphics[width=0.9\linewidth]{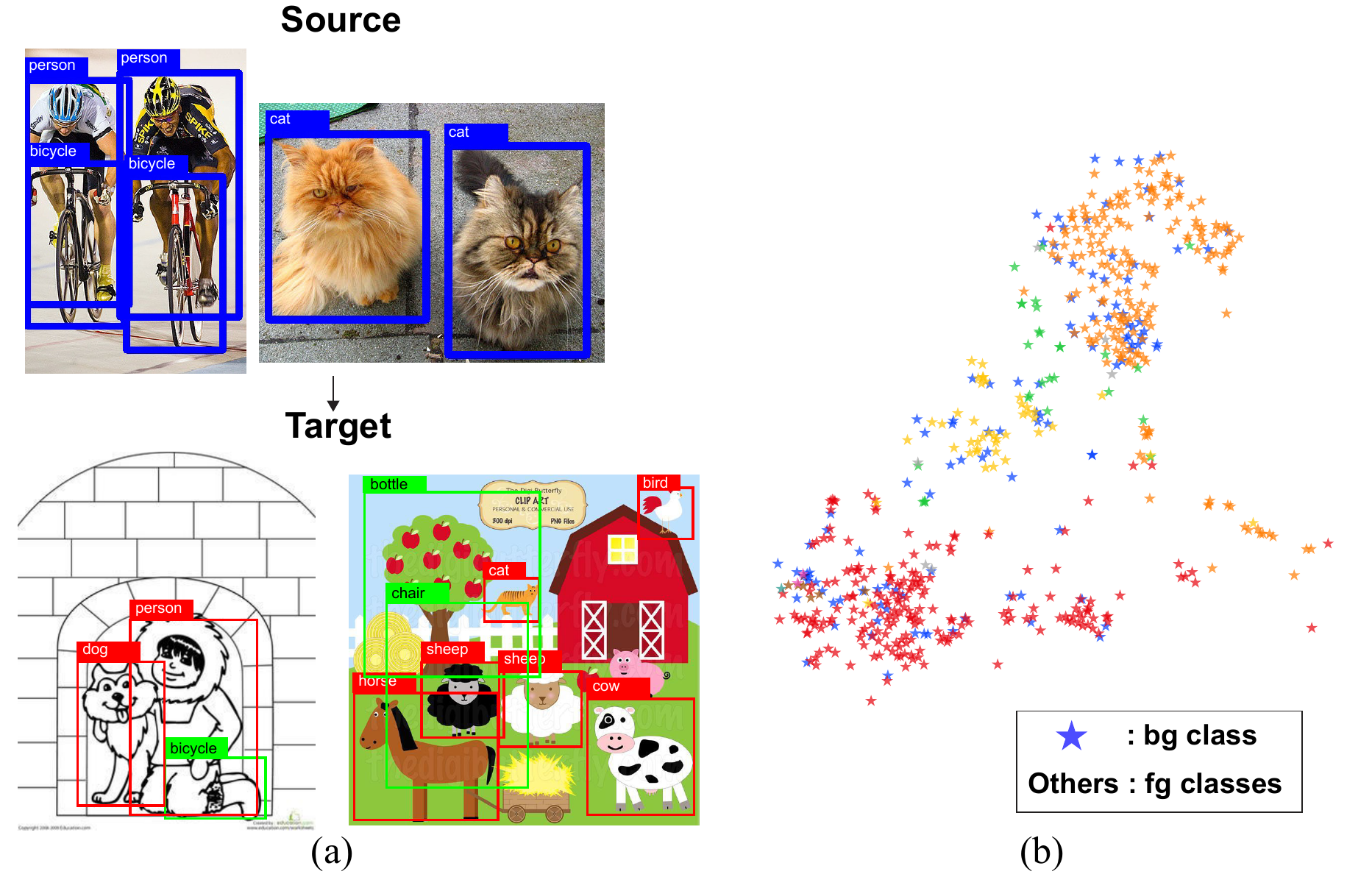}
\end{center}
\vspace{-10pt}
   \caption{(a) Due to domain discrepancy, the detector trained on the source domain does not perform well on the target. Green boxes indicate false positives and red indicate missing objects. (b) Feature visualization of the detection results on target images generated by source-only model. It is difficult to align feature at instance-level without category information due to the existence of false detections on background.}
\label{fig:problem definition}
\vspace{-10pt}
\end{figure}

\begin{figure*}[t]
\begin{center}
   \includegraphics[width=0.7\linewidth]{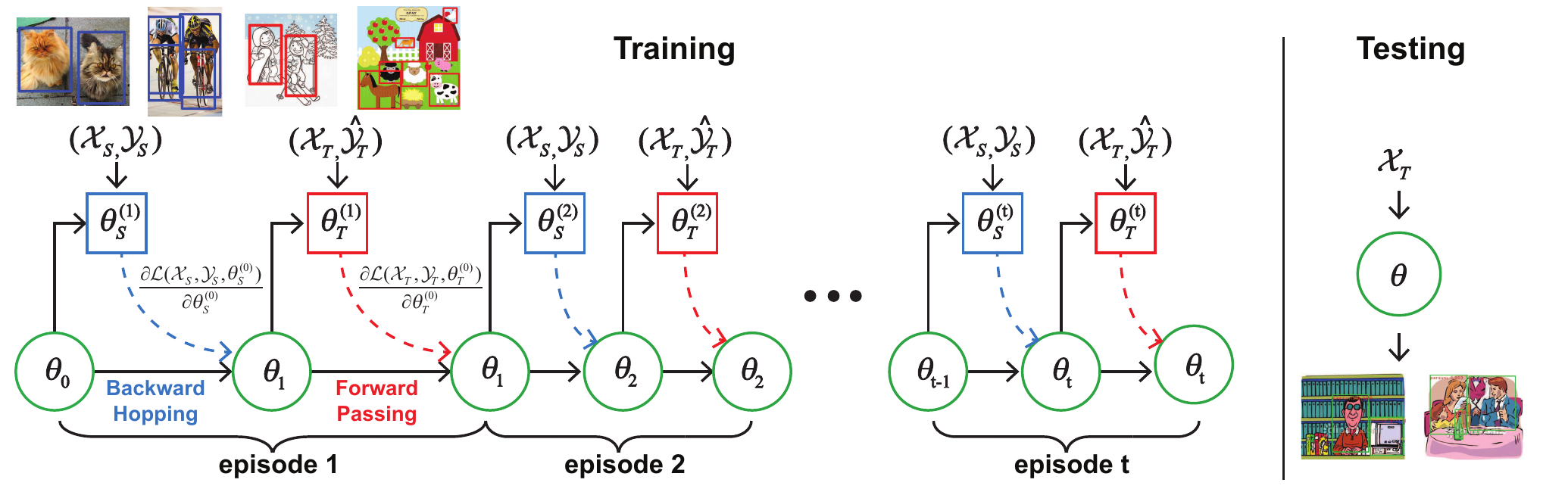}
\end{center}
\vspace{-10pt}
   \caption{The diagram of the proposed forward-backward cyclic adaptation for unsupervised domain adaptive object detection. In each episode, the training proceeds to achieve two goals: gradient alignment across the source $\mathcal{X}_s$ and target $\mathcal{X}_t$ to achieve domain invariant detectors; and encouraging domain-diversity to boost the target detector performance. }
\label{fig:diagram}
\end{figure*}


Object detection is a fundamental problem in computer vision \cite{fasterRCNN,SSD,YOLO,FCN,FPN}. It can be applied into many scenarios such as face and pedestrian detection~\cite{Tinyfaces} and self-driving cars~\cite{chen2017multi}. However, due to the variations in shape and appearance, lighting conditions and backgrounds, a model trained on the source data might not perform well on target--a problem known as \textit{domain discrepancy}, as shown in Figure~\ref{fig:problem definition}. A common approach to maximize the performance on the target domain is via fine-tuning a pre-trained model with a large amount of target data. However, the task of annotating bounding boxes for target objects is time-consuming and expensive. Hence, an effective model that can adapt object detectors into a new domain without labels--unsupervised domain adaptation--is highly desirable.

Unsupervised domain adaptation methods \cite{MMDlong2015,MMDlong2016,GRL,ADDA,AsymTriTraining,maximumClassifierDiscrepancy} for image classification are developed to learn domain-invariant features by minimizing the source error and simultaneously the domain discrepancy through feature distribution alignment. Standard optimization criteria include maximum mean discrepancy~\cite{MMDlong2015,MMDlong2016} and distribution moment matching~\cite{coral,tzeng2015simultaneous}. 
Recent adversarial training based domain adaptation methods have shown their effectiveness in learning domain-invariance by matching the marginal distributions of both source and target features \cite{GRL,DeepDomainConfusion,ADDA}. 
However, matching the marginal distributions does not guarantee the alignment of class conditional distributions~\cite{MSTN,DIRT-T,categoryLevelAdversaries,CoRegularizeAlignment}.
For example, aligning the target cat class to the source dog class can easily meet the objective of reducing the cost of source/target domain distinction, but the semantic categories are wrong.
The limitation of global adversarial learning is more evident when the domain discrepancy between two domains is larger.

In object detection, performing domain alignment is more challenging compared to alignment in the image classification task in the following two aspects: 
(1) The input image may contain multiple objects, while there is only one centered object in the classification task; 
(2) The images in object detection are dominated by background and non-objects.
Therefore, performing global adversarial learning (\ie, marginal feature distributions) at image-level is not sufficient for such a challenging task due to the limitations discussed above.
Chen~\etal\cite{adaptiveFasterRCNN} made the first attempt to apply domain adversarial training to object detection, where the marginal feature distributions at instance-level are aligned in addition to the image-level adaptation. 
However, due to the domain shift, the detector may not be accurate and
many non-object proposals from the backgrounds are used for domain alignment (Figure~\ref{fig:problem definition}).
This amplifies the limitation of domain adversarial training and hence limited gains can be achieved.
In order to tackle the limitation of global adversarial training, two state-of-the-art methods~\cite{adaptingObjSelective,SWDA} have been proposed to reduce the discrepancy between the aligned samples which are then used for adversarial training.
More specifically, they re-weigh and select the target images or proposals according to the scores of domain classifier, where images/instances have features similar to those of source samples will be emphasized.
The above methods show that it is challenging to align marginal feature distributions in the object detection task, especially when no target labels are provided.


In this work, we argue that explicit feature distribution alignment is not a necessary condition to learn domain-invariance. Instead, we remark that \textit{domain-invariance} of category level semantics can be achieved by gradient alignment, where the inner product between the gradients on images from different domains is maximized.
Intuitively, if the inner product is positive, the gradients from different domains are pointing in similar directions.
This implies that taking a gradient step on one domain can improve the learning on another domain and vice versa.
In other words, the two learnings share similar information and therefore lead to domain-invariance.
More importantly, the gradients of the last fully connected layer can preserve class conditional information for different domains.
Therefore, gradient alignment shows its advantages on the challenging task, adapting object detectors.

To achieve the goals, we propose a Forward-Backward Cyclic Adaptation (FBC) approach to learn adaptive object detectors. 
In each cycle, the games of \textit{Forward Passing}, adaptation from source to target, and \textit{Backward Hopping}, adaptation from target to source, are played sequentially.
Each adaptation is a domain transfer, where the training is first initialized with the model trained on previous domain and then finetuned with the images in current domain.
We provide theoretical analysis to show that by computing the forward and backward adaptation sequentially via Stochastic Gradient Descent (SGD), gradient alignment can be achieved.
Our proposed approach is related to the cycle consistency utilized in machine translation~\cite{DualLearningTranslation}, image-to-image translation~\cite{cyclegan,dualgan} and unsupervised domain adaptation~\cite{CycleUDA} with a similar intuition that the mappings of an example transferred from source to target and then back to the source domain should have same results.
Different from these approaches, we do not strictly enforce a cycle consistency loss on the source domain and our proposed cyclic adaptation shares the same network architectures for both adaptations rather than two separate generators.

In addition to learning the domain-invariance of category level semantics, we leverage domain adversarial training to learn the domain-invariance of holistic color and textures via aligning low level features.
This domain adversarial training is conducted on the adaptation from source to target domain.

However, a detector with good generalization on both domains may not be the optimal solution for the target domain.  
To address this, we introduce \textit{domain-diversity} into the training objective to avoid overfitting on the source domain and encourage target-specific learning on the target domain.
In this work, we adopt two regularizers: (1) a maximum entropy regularizer on source domain and (2) a minimum entropy regularizer on target domain.
The overview of our model is shown in Fig.~\ref{fig:diagram}.

We conduct experiments on four domain shift scenarios and experimental results show the effectiveness of our proposed approach.
\noindent \textbf{Contributions:} (1) We propose a forward-backward cyclic adaptation approach to learn unsupervised domain adaptive object detectors through effective gradient alignment; (2) To achieve good performance on the target domain, we explicitly enforce domain-diversity via entropy regularization to further approximate the domain-invariant detectors closer to the optimal solution to the target space; (3) The proposed method is simple yet effective and can be applied to various architectures. 

%
\section{Related Work}




\noindent
\textbf{Object Detection.}
Existing deep object detection methods~\cite{FastRCNN,fasterRCNN,YOLO,SSD,FPN,FocalLoss} can be roughly grouped into two categories: two-stage and single-stage frameworks. A representative of two-stage framework is the Faster R-CNN proposed by Ren~\etal\cite{fasterRCNN}, which consists of two sub-networks: a region proposal network that generates region proposals and a R-CNN that classifies the categories of the proposals. Single-stage detectors, \eg SSD~\cite{SSD} and YOLO~\cite{YOLO}, have demonstrated high efficiency in object detection, where the networks perform object classification and localization simultaneously. Other methods like FPN~\cite{FPN} and RetinaNet~\cite{FocalLoss} propose to leverage a combination of features from different levels to improve the feature representations.

\noindent
\textbf{Unsupervised Domain Adaptation for Image Classification.}
Unsupervised domain adaptation approaches are proposed to address domain discrepancy with labeled source data and unlabeled target data.
A vast number of deep learning based works are presented for image classification \cite{BeyondSharingWeights,ADDA,DeepDomainConfusion,MMDlong2015,MMDlong2016}.
Many adaptation methods~\cite{MMDlong2015,MMDlong2016,coral,JointAdaptationNetworks,GRL,DeepDomainConfusion,ADDA} are proposed to reduce the domain divergence based on the following theory:

\begin{theorem}[Ben-David~\etal\cite{Bn-David2010}]
Let $h: \mathcal{X} \to \mathcal{Y}$ be a hypothesis in the hypothesis space $\mathcal{H}$. The expected error on the target domain $\epsilon_T(h)$ is bounded by
\setlength{\belowdisplayskip}{3pt}
\setlength{\abovedisplayskip}{3pt}
\begin{equation}
\epsilon_T(h) \le \epsilon_S(h)+\frac{1}{2}d_{\mathcal{H}\Delta\mathcal{H}}(\mathcal{D}_S, \mathcal{D}_T) + \lambda, \forall h \in \mathcal{H} \textrm{ ,}
\label{eq: theorem 1}
\end{equation}
where $\epsilon_S(h)$ is the expected error on the source domain, 
\scalebox{0.8}{$d_{\mathcal{H}\Delta\mathcal{H}}(\mathcal{D}_S, \mathcal{D}_T) = 2\underset{h, h' \in \mathcal{H}}{\operatorname{sup}} \left|\underset{x \sim \mathcal{D}_S}{\operatorname{Pr}}[h(x) \neq h'(x)] - \underset{x \sim \mathcal{D}_T}{\operatorname{Pr}}[h(x) \neq h'(x)]\right|$}
measures domain divergence, and $\lambda$ is the expected error of ideal joint hypothesis, 
\scalebox{0.9}{$\lambda = \min_{h \in \mathcal{H}}[\epsilon_S(h)+\epsilon_T(h)]$}.
\label{theorem:ben david}
\end{theorem}
To minimize the divergence, various methods have been proposed to align the distributions of features from source and target domains,~\eg, maximum mean discrepancy~\cite{MMDlong2015,MMDlong2016}, correlation alignment~\cite{coral}, joint distribution discrepancy loss~\cite{JointAdaptationNetworks} and adversarial training that aligns marginal distributions~\cite{GRL,DeepDomainConfusion,ADDA}.
Recent adversarial training based methods \cite{GRL,DeepDomainConfusion,ADDA} are studied to match the marginal distributions of the source and target features, where the feature generator is trained to confuse the domain classifier.
Although the adversarial training based methods have achieved impressive results, recent works~\cite{MSTN,SWDA,DIRT-T,kang2019contrastive,chen2018progressive} show that aligning the marginal distributions without considering class conditional distributions does not guarantee small $d_{\mathcal{H}\Delta\mathcal{H}}(\mathcal{D}_S, \mathcal{D}_T)$. 
To address this, Luo~\etal\cite{categoryLevelAdversaries} propose to improve it via a semantic-aware discriminator, and Xie~\etal\cite{MSTN} propose to align the semantic prototypes for each class. 
Some works~\cite{MSTN,kang2019contrastive,chen2018progressive} propose to minimize the joint hypothesis error $\lambda$ with pseudo labels in addition to the marginal distribution alignment.
Some other methods propose to use the predictions of a classifier as pseudo labels for unlabeled target samples~\cite{CycleUDA,AsymTriTraining,CoTraining}. 
Lee~\etal\cite{Pseudo-label} argue that training with pseudo labels is equivalent to entropy regularization, which favors a low density separation between classes.

%
%
%
%
%
%
%


\noindent
\textbf{Unsupervised Domain Adaptation for Object Detection.}
Fewer works are available in the unsupervised domain adaptation for object detection. To our knowledge, there are only three works, Domain Adaptive Faster R-CNN (DA-Faster)~\cite{adaptiveFasterRCNN}, Selective Domain Alignment (SDA)~\cite{adaptingObjSelective} and Strong-Weak Domain Alignment (SWDA)~\cite{SWDA}. The DA-Faster~\cite{adaptiveFasterRCNN} adds two domain classifiers to the Faster-RCNN for learning domain-invariant features for image-level and instance-level features. 
However, due to the limitation of domain adversarial training and inaccurate instance predictions, the improvement is limited.
To address this, two state-of-the-art methods~\cite{adaptingObjSelective,SWDA} propose to select target images/instances that are similar to source ones for adversarial training.
Zhu~\etal\cite{adaptingObjSelective} propose to first filter non-objects via grouping the proposals and then emphasize the target proposals that are similar to the source for adversarial domain alignment.
Saito~\etal\cite{SWDA} propose to weakly align the image-level features from the high-level layer, where the images that are globally similar have higher priority to be aligned.
The weak alignment is achieved by replacing the cross-entropy loss of domain classifier with focal loss~\cite{FocalLoss}.
In contrast to selecting similar pairs for adversarial training, our proposed method learns domain-invariance of category level semantics via gradient alignment. 

%

\noindent
\textbf{Gradient-based Meta Learning.}
Our method is also related to recent gradient-based meta learning methods: MAML~\cite{MAML} and Reptile~\cite{reptile}, which are designed to learn a good initialization for few shot learning and have demonstrated good within-task generalization. 
Reptile~\cite{reptile} suggested that SGD automatically maximizes the inner products between the gradients computed on different minibatches of the same task, and results in within-task generalization.
Riemer~\etal~\cite{MaxTransferMinInterference} integrates the Reptile algorithm with an experience replay module for the task of continual learning, where the transfer between examples is maximized via meta-learning.
Inspired by these methods, we leverage the generalization ability of Reptile~\cite{reptile} to improve the generalization across domains for unsupervised domain adaptation via gradient alignment.


%
\section{Forward-Backward Domain Adaptation for Object Detection}
\subsection{Overview}
In unsupervised domain adaptation, $N_S$ labeled images $\{\mathcal{X}_S, \mathcal{Y}_S\}=\{x^i_S, y^i_S\}^{N_S}_{i=1}$ from the source domain with a distribution $\mathcal{D}_S$ are given. We have $N_T$ unlabeled images $\mathcal{X}_T=\{x^j_T\}^{N_T}_{j=1}$ from the target domain with a different distribution $\mathcal{D}_T$, but the ground truth labels $\mathcal{Y}_T=\{y^j_T\}^{N_T}_{j=1}$ are not accessible during training. Note that in object detection, each label in $\mathcal{Y}_S$ or $\mathcal{Y}_T$ is composed of a set of bounding boxes with their corresponding class labels. Our goal is to learn a neural network (parameterized by $\theta$) $f_{\theta}: \mathcal{X}_T \to \mathcal{Y}_T$ that can make accurate predictions on the target samples without the need for labeled training data. 



In Theorem~\ref{theorem:ben david}, the expected error on target domain $\epsilon_T(h)$ is bounded by three terms: (1) the expected error on source domain $\epsilon_S(h)$ which can be minimized easily via supervised learning, (2) the disagreement between two hypotheses on source and target domains $d_{\mathcal{H}\Delta\mathcal{H}}(\mathcal{D}_S, \mathcal{D}_T)$, and (3) the expected error of ideal joint hypothesis $\lambda$. 
In this work, we argue that aligning the feature distributions is not a necessary condition to reduce the $d_{\mathcal{H}\Delta\mathcal{H}}(\mathcal{D}_S, \mathcal{D}_T)$.
Different from the above-mentioned distribution alignment based methods,  we cast the domain adaptation into an optimization problem to leverage the domain-invariance.
As the ultimate goal of domain adaptation is to achieve good performance on the target domain, we further introduce \textit{domain-diversity} into training to boost the detection performance in the target space.

\subsection{Gradient Alignment via Forward-Backward Cyclic Training}

\begin{figure}[t]
\begin{center}
   \includegraphics[width=0.7\linewidth]{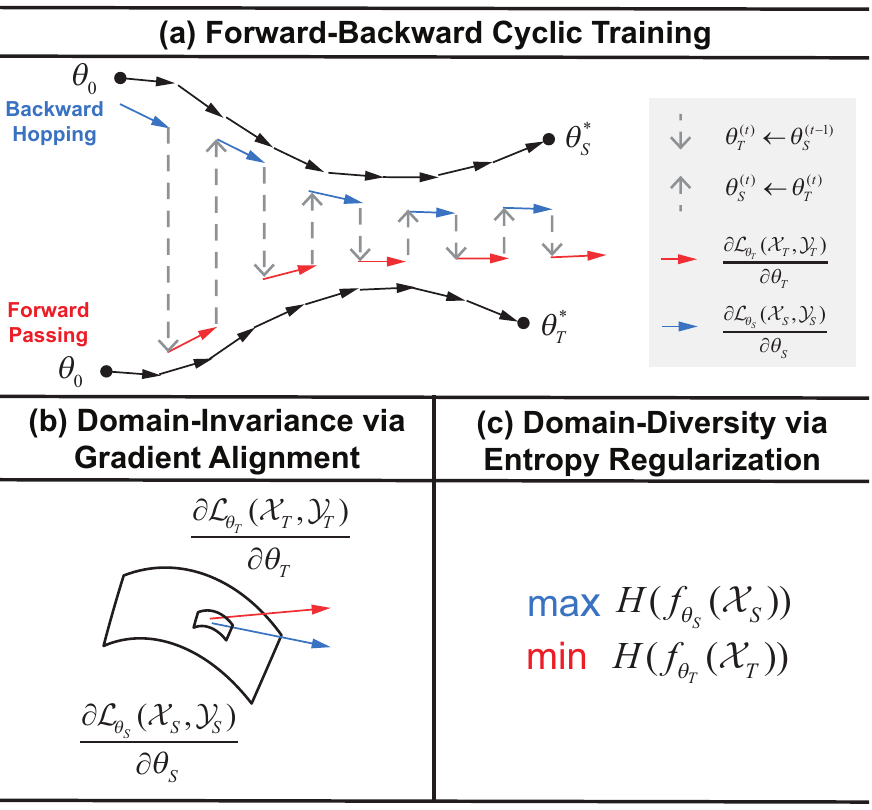}
\end{center}
\vspace{-10pt}
   \caption{(a) Illustration of the model updates in our proposed forward-backward cyclic adaptation method. The $\theta_0$ is the initial model and the $\theta^*_S$ and $\theta^*_T$ are the optimal solutions for source and target domain respectively. (b) We propose that domain-invariance occurs when the gradients of source and target samples are pointing in similar directions. (c) The domain diversity, which is introduced to avoid overfitting on source domain.}
\label{fig:invariance}
\end{figure}



Recent gradient-based meta-learning methods~\cite{MAML,optimizationMetaLearning,reptile}, designed for few shot learning, have demonstrated their success in approximating learning algorithms and shown their ability to generalize well to new data from unseen distributions. 
Inspired by these methods, we propose to learn the \textit{domain-invariance} via gradient alignment to achieve generalization across domains.

\subsubsection{Gradient Alignment for Domain-invariance}

Suppose that we have neural networks that learn the predictions for source and target samples as $f_{\theta_S}: \mathcal{X}_S \to \mathcal{Y}_S$ and $f_{\theta_T}: \mathcal{X}_T \to \mathcal{Y}_T$, respectively. The network parameters $\theta_S$ and $\theta_T$ are updated via minimizing the empirical risks, $\mathcal{L}_{\theta_S}(\mathcal{X}_S, \mathcal{Y}_S)=\frac{1}{N_S}\sum_{i=1}^{N_S}\ell(f_{\theta_S}(x^i_S), y^i_S)$ and $\mathcal{L}_{\theta_T}(\mathcal{X}_T, \mathcal{Y}_T)=\frac{1}{N_T}\sum_{j=1}^{N_T}\ell(f_{\theta_T}(x^j_T), y^j_T)$, where $\ell(\cdot)$ is the cross-entropy loss.
%
%
%
%
%
%


In this paper, we argue that domain-invariance occurs when:
\setlength{\belowdisplayskip}{5pt}
\setlength{\abovedisplayskip}{5pt}
\begin{equation}
\frac{\partial \mathcal{L}_{\theta_S}(\mathcal{X}_S, \mathcal{Y}_S)}{\partial \theta_S} \cdot \frac{\partial \mathcal{L}_{\theta_T}(\mathcal{X}_T, \mathcal{Y}_T)}{\partial \theta_T} > 0 \textrm{ ,}
\label{eq:domain invariance}
\end{equation}
where the $\cdot$ is the inner-product operator. When two gradients are pointing in similar directions, it implies that the learning of source samples can benefit the learning of target samples and vice versa. This indicates that the two learnings share similar information and therefore leads to domain-invariance. Moreover, this gradient alignment can encode conditional class information as the gradients are generated from the classification losses $\mathcal{L}_{\theta_S}(\mathcal{X}_S, \mathcal{Y}_S)$ and $\mathcal{L}_{\theta_T}(\mathcal{X}_T, \mathcal{Y}_T)$.
This is different from the feature alignment by a domain classifier in adversarial training based methods~\cite{GRL,DeepDomainConfusion,ADDA,adaptiveFasterRCNN,SWDA}, where class information is not explicitly considered.

Recall Theorem~\ref{eq: theorem 1}, once $d_{\mathcal{H}\Delta\mathcal{H}}(\mathcal{D}_S, \mathcal{D}_T)$ is minimized, the generalization error on target domain $\epsilon_T(h)$ is bounded by the shared error of ideal joint hypothesis, $\lambda = \min_{h \in \mathcal{H}}[\epsilon_S(h)+\epsilon_T(h)]$. 
As suggested in \cite{Bn-David2010}, it is important to have a classifier performing well on both domains. 
Therefore similar to the previous works~\cite{Pseudo-label,MSTN,chen2018progressive}, we resort to using pseudo labels $\mathcal{\hat{Y}}_T=\{\hat{y}^j_T\}^{N_T}_{j=1}$ to optimize the upper bound for the $\lambda$.
These pseudo labels are the detections on the target images produced by the source detector $f_{\theta_S}$ and are updated with the updates of $f_{\theta_S}$.
Therefore, our objective function of gradient alignment is
\setlength{\belowdisplayskip}{5pt}
\setlength{\abovedisplayskip}{5pt}
\begin{align}
\min_{\theta_S, \theta_T}  & \mathcal{L}_{g}(\mathcal{X}_S, \mathcal{Y}_S, \mathcal{X}_T, \mathcal{\hat{Y}}_T) \notag\\
&
= \mathcal{L}_{\theta_S}(\mathcal{X}_S, \mathcal{Y}_S)+\mathcal{L}_{\theta_T}(\mathcal{X}_T, \mathcal{\hat{Y}}_T) \notag\\
& 
\quad - \alpha \frac{\partial \mathcal{L}_{\theta_S}(\mathcal{X}_S, \mathcal{Y}_S)}{\partial \theta_S} \cdot \frac{\partial \mathcal{L}_{\theta_T}(\mathcal{X}_T, \mathcal{\hat{Y}}_T)}{\partial \theta_T}
\textrm{ .}
\label{eq:objective function gradient alignment}
\end{align}

\subsubsection{Forward-Backward Cyclic Training}

To achieve the above objective, we propose an algorithm that sequentially plays the game of \textit{Backward Hopping} on the source domain and \textit{Foward Passing} on the target domain, and a shared network parameterized by $\theta$ is updated iteratively. We initialize the shared network $\theta$ with ImageNet~\cite{imagenet} pre-trained model. Let us denote a cycle of performing forward passing and backward hopping as an episode. 
In the backward hopping phase of episode $t$, the network parameterized $\theta_S^{(t)}$ is first initialized with the model $\theta_T^{(t-1)}$ from the previous episode $t-1$.
And the model $\theta_S^{(t)}$ is then optimized with one image per time via stochastic gradient descent (SGD) on $N_S$ labeled source images $\{\mathcal{X}_S, \mathcal{Y}_S\}$. 
In forward passing, the model $\theta_T^{(t)}$ is initialized with $\theta_S^{(t)}$ and trained with pseudo labeled target samples $\{\mathcal{X}_T, \hat{\mathcal{Y}}_T\}$.
The training procedure is shown in Fig.~\ref{fig:diagram}.

\medskip
\noindent
\textbf{Theoretical Analysis.}
We provide theoretical analysis to show how our proposed forward and backward training strategy can achieve the objective of gradient alignment in Eq.~\ref{eq:objective function gradient alignment}.
For simplicity, we only analyze the gradient computations in one episode and denote the gradient obtained in one episode as $g_{e}$.
We then have $g_{e}=g_S+g_T$, where 
$g_S$ is obtained in backward hopping
$g_S = \frac{\partial \mathcal{L}_{\theta_S}(\mathcal{X}_S, \mathcal{Y}_S)}{\partial \theta_S}$
and $g_T$ is the gradient obtained in forward passing
$g_T = \frac{\partial \mathcal{L}_{\theta_T}(\mathcal{X}_T, \mathcal{\hat{Y}}_T)}{\partial \theta_T}$.

According to the Taylor's theorem, the gradient of forward passing can be expanded as
$g_T=\bar{g}_T+\bar{H}_T(\theta_T-\theta_0) + O(\lVert \theta_T - \theta_0 \rVert ^2)$, where $\bar{g}_T$ and $\bar{H}_T$ are the gradient and Hessian matrix at initial point $\theta_0$.
Then the overall gradient $g_{e}$ can be rewritten as:
\begin{align}
g_{e} & =g_S+g_T \notag\\
&=\bar{g}_S+\bar{g}_T+\bar{H}_T(\theta_T-\theta_0) + O(\lVert \theta_T - \theta_0 \rVert ^2) \textrm{ .}
\label{eq:g gradient}
\end{align}

Let us denote the initial parameters in one episode as $\theta_0$.
In our proposed forward and backward training strategy, the model parameters of backward hopping are first initialized with $\theta_S=\theta_0 $ and are updated by $\theta_0 - \alpha g_S$.
In forward passing, the model is initialized with the updated $\theta_S$ and thus $\theta_T=\theta_0 - \alpha g_S$.
Substitute this to Eq.~\ref{eq:g gradient} and we have
\begin{align}
g_{e}=\bar{g}_S+\bar{g}_T - \alpha\bar{H}_T \bar{g}_S + O(\lVert \theta_T - \theta_0 \rVert ^2) \textrm{ .}
\label{eq:g gradient2}
\end{align}

It is noted in Reptile~\cite{reptile} that $\mathbb{E}[\bar{H}_S\bar{g}_T]=\mathbb{E}[\bar{H}_T\bar{g}_S]=\frac{1}{2}[\frac{\partial}{\partial \theta_0}(\bar{g}_S \cdot \bar{g}_T)]$.
Therefore, this training is approximating our objective function in Eq.~\ref{eq:objective function gradient alignment}. The proposed training strategy is relevant to the meta-learning approaches~\cite{MAML,reptile,MLDG} that are initially designed for few shot learning. Their training mainly aims at the generalization ability within one task, while our forward and backward training is aiming at the generalization across tasks.
More details are shown in the supplementary materials.

\subsection{Local Feature Alignment via Adversarial Training}

\begin{figure}[t]
\begin{center}
   \includegraphics[width=1\linewidth]{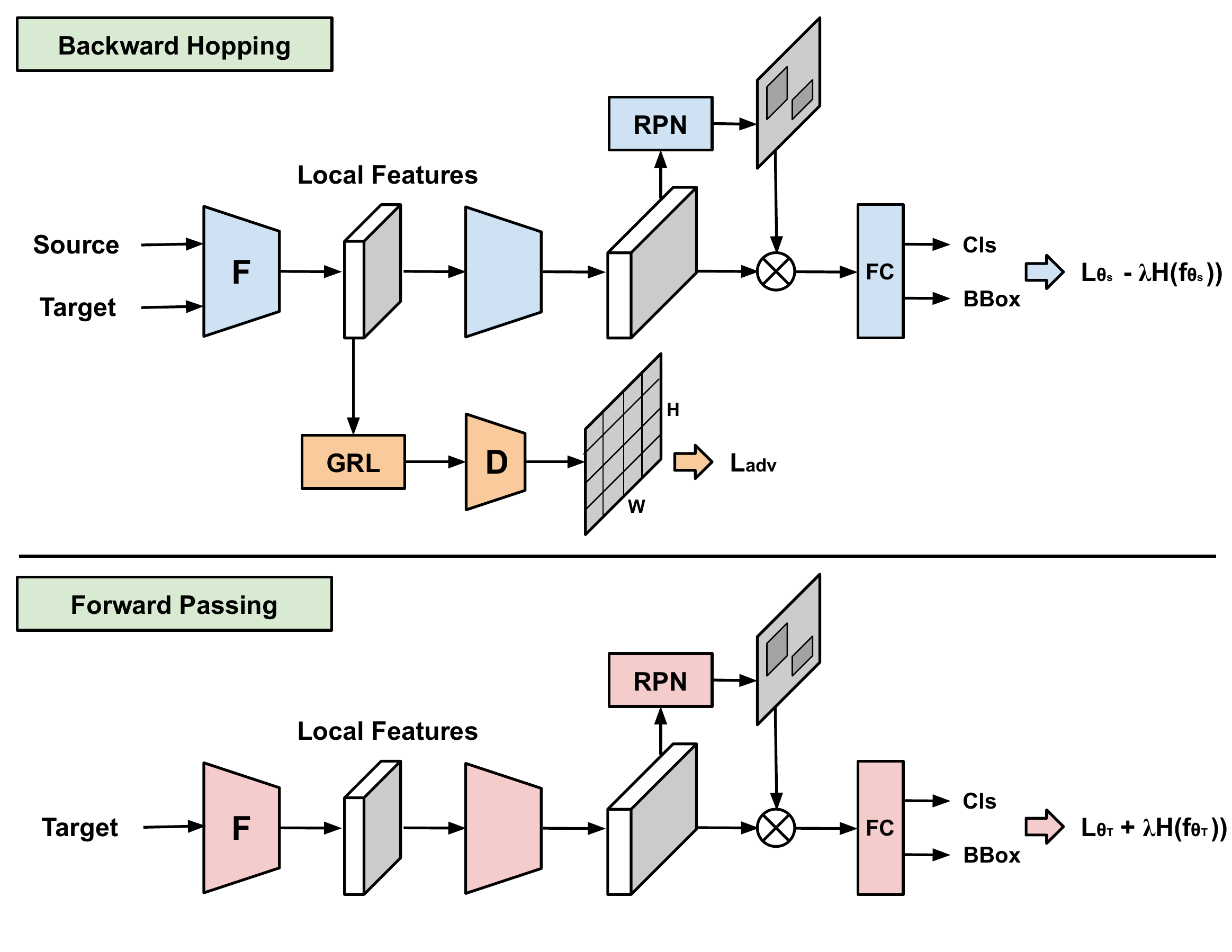}
\end{center}
\vspace{-10pt}
   \caption{Network Architecture.}
\label{fig:invariance}
\vspace{-10pt}
\end{figure}

Domain adversarial training has demonstrated its effectiveness in reducing domain discrepancy of low-level features, \eg local texture and color, regardless of class conditional information~\cite{adaptiveFasterRCNN,SWDA}.
Therefore, we align the low-level features at image-level in combination with the gradient alignment on source domain.
We utilize the gradient reversal layer (GRL) proposed by Ganin and Lempitsky~\cite{GRL} for domain adversarial training, where the gradients of the domain classifier are reversed for domain confusion.
Following SWDA~\cite{SWDA}, we extract local features $F$ from low-level layer as input of the domain classifier $D$
and the least-squares loss~\cite{LSGAN,cyclegan} is used to optimize the domain classifier.
The loss of adversarial training is as follows:
\begin{align}
\mathcal{L}_{adv} = &\frac{1}{2}\frac{1}{N_SWH}\sum_{i, w, h}^{} D(F(x_S^i))_{wh}^2 \notag\\
& + \frac{1}{2}\frac{1}{N_SWH}\sum_{j, w, h}^{} (1-D(F(x_T^j))_{wh})^2 \textrm{ ,}
\end{align}
where $H$ and $W$ are height and width of the output feature map of domain classifier.

\subsection{Domain Diversity via Entropy Regularization}
The ultimate goal of domain adaptation is to achieve good performance on target domain.
However, a model that only learns the domain-invariance is not an optimal solution for the target domain, as 
\begin{align}
\epsilon_T(h) & \le\epsilon_T(h^a)+\epsilon_T(h, h^a) 
,
\end{align}
where $h^a = \argmin_{h \in \mathcal{H}} [\epsilon_S(h) + \epsilon_T(h)]$.
Moreover, in the absence of ground truth labels for target samples, the learning of domain-invariance largely relies on the source samples, which might results in the overfitting on the source domain and limits its ability to generalize well on target domain.
Therefore, it is important to introduce the domain-diversity into the training to encourage more emphasis on target-specific information.

In this work, we define the domain diversity as a combination of two regularizations: (1) maximum entropy regularization on source domain to avoid overfitting and (2) minimum entropy regularization on unlabled target domain to leverage target-specific information.
Low entropy corresponds to high confidence.
To avoid the overfitting when training with source domain data, we utilize the maximum entropy regularizer~\cite{pereyra2017regularizing}, which penalizes the confident predictions with low entropy.
The maximum entropy principle proposed by Jaynes~\cite{jaynes1957information} has been applied to reinforcement learning~\cite{williams1991function,mnih2016asynchronous} to prevent early convergence and supervised learning to improve generalization~\cite{pereyra2017regularizing,temporal_cls_maximum_entropy,finegrained_maximum_entropy}.
\begin{equation}
\max_{\theta_S} \textrm{H}(f_{\theta_S}(\mathcal{X}_S))=-\sum_{i=1}^{N_S} f_{\theta_S}(x^i_S)\operatorname{log}(f_{\theta_S}(x^i_S))\textrm{ .}
\label{eq:diversity max entropy}
\end{equation}
On the contrary, to leverage unlabeled target domain data, we exploit the minimum entropy regularizer.
This entropy minimization has been used for unsupervised clustering~\cite{clustering_entropy_minimization}, semi-supervised learning~\cite{SSL_entropy_minimization} and unsupervised domain adaptation~\cite{MMDlong2016,zelunluo2017label} to encourages low density separation between clusters or classes.
Here, we minimize the entropy of class conditional distribution:
\begin{equation}
\min_{\theta_T} \textrm{H}(f_{\theta_T}(\mathcal{X}_T))=-\sum_{j=1}^{N_T} f_{\theta_T}(x^j_T)\operatorname{log}(f_{\theta_T}(x^j_T))\textrm{ .}
\label{eq:diversity max entropy}
\end{equation}
We define the objective function of learning domain diversity as:
\begin{equation}
\mathcal{L}_{div}(\mathcal{X}_S, \mathcal{X}_T)=-\textrm{H}(f_{\theta_S}(\mathcal{X}_S))+\textrm{H}(f_{\theta_T}(\mathcal{X}_T))\textrm{ .}
\label{eq:domain diversity}
\end{equation}

%
%
%
\subsection{Overall Objective}
To learn domain-invariance for adapting object detectors, we propose to perform gradient alignment for high-level semantics and domain adversarial training on local features for low-level information, \eg local textures/color. 
\begin{align}
\mathcal{L}_{inv}(\mathcal{X}_S, \mathcal{Y}_S, \mathcal{X}_T)= \mathcal{L}_g(\mathcal{X}_S, \mathcal{Y}_S, \mathcal{X}_T) + \lambda\mathcal{L}_{adv}(\mathcal{X}_S,  \mathcal{X}_T)\textrm{ ,}
\label{eq:overall invariance}
\end{align}
where $\lambda$ balances the trade-off between gradient alignment loss and adversarial training loss.

Maximizing the domain-diversity contradicts the intention of learning domain-invariance.
As discussed above, without the access to the ground truth labels of target samples, the accuracy on the target samples relies on the domain-invariance information learnt from the source domain.
Consequently, to accomplish the trade-off between learning domain-invariance and domain-diversity, we propose to use a hyperparameter $\gamma$ to balance the trade-off.
Therefore, our overall objective function is 
\begin{align}
\min_{\theta} \mathcal{L}_{inv}(\mathcal{X}_S, \mathcal{Y}_S, \mathcal{X}_T) + \gamma \mathcal{L}_{div}(\mathcal{X}_S, \mathcal{X}_T)\textrm{ .}
\label{eq:objective function}
\end{align}
The full algorithm is outlined in Algorithm~\ref{alg1}.

\begin{algorithm}[t] 
\small
\caption{Forward-Backward Cyclic Domain Adaptation for Object Detection} 
\label{alg1} 
\begin{algorithmic}[1] 
    \REQUIRE Source samples $\{x^i_S, y^i_S\}^{N_S}_{i=1}$, target samples $\{x^j_T\}^{N_T}_{j=1}$, ImageNet pre-trained model $\theta_0$, hyperparameters $\alpha$, $\beta$, $\gamma$, $\lambda$, number of iterations $N_{itr}$
    \ENSURE A shared model $\theta$
    \STATE Initialize $\theta$ with $\theta_0$
    
    \FOR{$t$ in $N_{itr}$}
        \STATE \textit{\textbf{//Backward Hopping:}}
        \STATE$\theta_S^{(t)} \gets \theta$
        \FOR{$i$, $j$ in $N_{S}$, $N_T$}
            \STATE $\theta_S^{(t)} \gets \theta_S^{(t)}  - \alpha (\nabla_{\theta_S^{(t)}}(\mathcal{L}_{\theta_S^{(t)}}(x^i_S, y^i_S) +\lambda \mathcal{L}_{adv}(x^i_S, x^j_T))-\gamma \textrm{H}(f_{\theta_S^{(t)}}(x^i_S)))$
        \ENDFOR   
        
        \STATE$\theta \gets \theta - \beta\theta_S^{(t)}$      
        \STATE Generate pseudo labels $\hat{y}_T = f_{\theta_S^{(t)}}(x^j_T), j=1,...,N_T$
		\STATE \textit{\textbf{//Forward Passing:}}
			\STATE$\theta_T^{(t)} \gets \theta$
        	\FOR{$j$ in $N_{T}$}
            	\STATE $\theta_T^{(t)} \gets \theta_T^{(t)} - \alpha ( \nabla_{\theta_T^{(t)}} (\mathcal{L}_{\theta_T^{(t)}}(x^j_T, \hat{y}^j_T)+\gamma \textrm{H}(f_{\theta_T^{(t)}}(x^j_T)))$
        	\ENDFOR
        
        	\STATE$\theta \gets \theta - \beta \theta_T^{(t)}$

    \ENDFOR
\end{algorithmic}
\end{algorithm}
\vspace{-5pt}
\section{Experiments}
In this section, we evaluate the proposed forward-backward cycling adaptation approach (FBC) on four cross-domain detection datasets. 

\subsection{Implementation Details}


Following DA-Faster~\cite{adaptiveFasterRCNN} and SWDA~\cite{SWDA}, we use the Faster-RCNN~\cite{fasterRCNN} as our detection framework. All training and test images are resized with the shorter side of 600 pixels and training batch size is 1.
Our method is implemented using Pytorch.

\noindent
\textbf{Baselines.} 
We compare our method with three baselines: 
(1) \textit{Source Only}: 
A Faster R-CNN detector that is fine-tuned on the pre-trained ImageNet~\cite{imagenet} model with labeled source samples without adaptation;
%
%
(2) \textit{DA-Faster}~\cite{adaptiveFasterRCNN};
%
(3) \textit{SWDA}~\cite{SWDA};
(4) \textit{Zhu}~\etal~\cite{adaptingObjSelective}.

%
\noindent
\textbf{Evaluation Metrics.}
For the evaluation, we measure the mean average precision (mAP) with a threshold of 0.5 across all classes.
As the source only models are a bit different among baselines, the gain of mAP is used as a metric to evaluate the effectiveness of adaptation.


\subsection{Adaptation between Dissimilar Domains}
\label{subsec: results clipart and watercolor}

\begin{table*}[ht]
\scriptsize
\begin{center}
\begin{tabular}{l|p{0.2cm}p{0.22cm}*{17}{p{0.21cm}}p{0.29cm}|p{0.4cm}}
\hline
Method &aero &bcycle &bird &boat &bottle &bus &car &cat &chair &cow &table &dog &hrs &motor &prsn &plnt &sheep &sofa &train &tv & mAP \\
\hline
Source Only~\cite{SWDA} &35.6 &52.5 &24.3 &23.0 &20.0 &43.9 &32.8 &10.7 &30.6 &11.7 &13.8 &6.0 &36.8 &45.9 &48.7 &41.9 &16.5 &7.3 &22.9 &32.0 &27.8  \\
DA-Faster\dag~\cite{adaptiveFasterRCNN} &15.0 &34.6 &12.4 &11.9 &19.8 &21.1 &23.2 &3.1 &22.1 &26.3 &10.6 &10.0 &19.6 &39.4  &34.6 &29.3 &1.0 &17.1 &19.7 &24.8 &19.8 \\
SWDA~\cite{SWDA} &26.2 &48.5 &32.6 &33.7 &38.5 &54.3 &37.1 &18.6 &34.8 &58.3 &17.0 &12.5 &33.8 &65.5 &61.6 &52.0 &9.3 &24.9 &54.1 &49.1 &38.1  \\
\hline
Source Only (ours) &24.2 &47.1 &24.9 &17.7 &26.6 &47.3 &30.4 &11.9 &36.8 &26.4 &10.1 &11.8 &25.9 &74.6 &42.1 &24.0 &3.8 &27.2 &37.9 &29.9 &29.5 \\
FBC (ours) &43.9 &64.4 &28.9 &26.3 &39.4 &58.9 &36.7 &14.8 &46.2 &39.2 &11.0 &11.0 &31.1 &77.1 &48.1 &36.1 &17.8 &35.2 &52.6 &50.5 &\textbf{38.5} \\
FBC w/o (ours) &32.1 &57.6 &24.4 &23.7 &34.1 &59.3 &32.2 &9.1 &40.3 &41.3 &27.8 &11.9 &30.2 &72.9 &48.8 &38.3 &6.1 &33.1 &46.5 &48.0 &36.0\\
\hline
\end{tabular}
\end{center}
\vspace{-5pt}
\caption{Results ($\%$) on the adaptation from PASCAL~\cite{PASCAL} to Clipart Dataset~\cite{clipart_watercolor}. The DA-Faster\dag result is the reported in SWDA~\cite{SWDA}.}
\label{tab:clipart}
\vspace{-5pt}
\end{table*}

We evaluate the adaptation performance on two pairs of dissimilar domains: PASCAL~\cite{PASCAL} to Clipart~\cite{clipart_watercolor}, and PASCAL \cite{PASCAL} to Watercolor~\cite{clipart_watercolor}. For the two domain shifts, we use the same source-only model trained on PASCAL. Following SWDA~\cite{SWDA}, we use ResNet101~\cite{ResNet} as the backbone network for Faster R-CNN detector and the settings of training and test sets are the same.

\noindent
\textbf{Datasets.} 
PASCAL VOC dataset~\cite{PASCAL} is used as source domain in these two domain shift scenarios.
This dataset consists of real images with 20 object classes.
The training set contains around 15K images.
The two dissimilar target domains are the Clipart dataset~\cite{clipart_watercolor} with comic images and the Watercolor dataset~\cite{clipart_watercolor} with artistic images.
Clipart dataset has the same 20 object classes as the PASCAL, while Watercolor only has six. Clipart dataset contains 1K comic images, which are used for both training (without labels) and testing. There are 2K images in the Watercolor dataset: 1K for training (without labels) and 1K for testing.

\noindent
\textbf{Results on Clipart Dataset~\cite{clipart_watercolor}.} 
In the original paper of DA-Faster~\cite{adaptiveFasterRCNN}, they do not evaluate on the Clipart and Watercolor datasets. Thus, we follow with the results of DA-Faster \cite{adaptiveFasterRCNN} reported in SWDA \cite{SWDA}.
As shown in Table~\ref{tab:clipart}, in comparison to the source only model, DA-Faster \cite{adaptiveFasterRCNN} degrades the detection performance significantly, with a drop of $8$ percentage points in mAP. DA-Faster \cite{adaptiveFasterRCNN}  adopts two domain classifiers on both image-level and instance-level features. However, the source/target domain confusion without considering the semantic information will lead to wrong alignment of semantic classes across domains.
The problem is more challenging when domain shift in object detection is large, \ie PASCAL~\cite{PASCAL} to Clipart~\cite{clipart_watercolor}.
In Clipart \cite{clipart_watercolor}, the comic images contain objects that are far different from those in PASCAL \cite{PASCAL} w.r.t. the shapes and appearance, such as sketches.  
To address this, the SWDA~\cite{SWDA} conducts a weak alignment on the image-level features by training the domain classifier with a focal loss.
With the additional help of a domain classifier on lower level features and context regularization,
the SWDA \cite{SWDA} can boost the mAP of detection from 27.8\% to 38.1\% with an increase of 10.3 points.
Our proposed FBC can achieve the highest mAP of 38.5\%.


\begin{figure}[t]
\begin{center}
   \includegraphics[width=\linewidth]{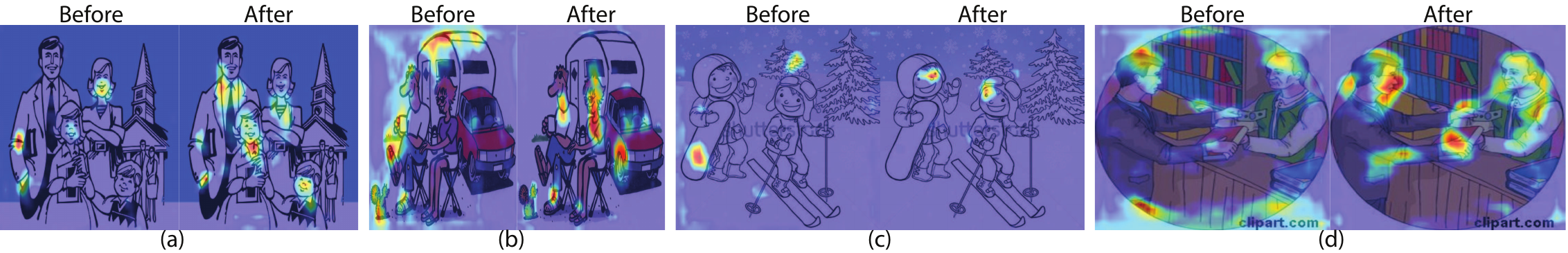}
\end{center}
\vspace{-10pt}
   \caption{Feature visualization for showing the evidence for classifiers before and after domain adaptation using Grad-cam~\cite{grad-cam}.}
   \label{fig:detection examples}
   \vspace{-10pt}
\end{figure}



\begin{table}[h]
\scriptsize
\begin{center}
\begin{tabular}{l|*{6}{p{0.24cm}}|p{0.3cm}}
\hline
Method  &bike &bird &car &cat &dog &prsn &mAP\\
\hline
Source Only~\cite{SWDA} &68.8 &46.8 &37.2 &32.7 &21.3 &60.7 &44.6 \\
DA-Faster\dag~\cite{adaptiveFasterRCNN}  &75.2 &40.6 &48.0 &31.5 &20.6 &60.0 &46.0 \\
SWDA~\cite{SWDA}  &82.3 &55.9 &46.5 &32.7 &35.5 &66.7 &53.3  \\
\hline
Source Only (ours) &66.7 &43.5 &41.0 &26.0 &22.9 &58.9 &43.2 \\
FBC (ours)  &90.9 &47.7 &46.0 &38.7 &31.8 & 66.7 &\textbf{53.6}  \\
FBC w/o local (ours) &88.7 &48.2 &46.6 &38.7 &35.6 &64.1 &\textbf{53.6} \\
\hline
\end{tabular}
\end{center}
\vspace{-5pt}
\caption{Results ($\%$) on the adaptation from PASCAL~\cite{PASCAL} to Watercolor~\cite{clipart_watercolor}. The DA-Faster\dag is the reproduced in SWDA~\cite{adaptiveFasterRCNN}.}
\label{tab:watercolor}
\end{table}
%
%
\begin{table}[h]
\scriptsize
\begin{center}
\begin{tabular}{l|c}
\hline
Method  &AP on Car \\
\hline
Source Only~\cite{adaptiveFasterRCNN} &31.2 \\
DA-Faster~\cite{adaptiveFasterRCNN}  &39.0  \\
\hline
Source Only~\cite{SWDA} &34.6 \\
DA-Faster\dag~\cite{adaptiveFasterRCNN}  &34.2  \\
SWDA~\cite{SWDA}  &42.3  \\
\hline
Source Only~\cite{adaptingObjSelective} &34.0 \\
Zhu~\etal\cite{adaptingObjSelective}  &\textbf{43.0}  \\
\hline
Source Only (ours) &31.2  \\
FBC (ours) &42.7 \\
FBC w/o local (ours) &39.2 \\
\hline
\end{tabular}
\end{center}
\vspace{-5pt}
\caption{Results ($\%$) on the adaptation from Sim10k~\cite{sim10k} to Cityscapes~\cite{cityscapes}. The DA-Faster\dag is the reproduced in SWDA~\cite{adaptiveFasterRCNN}.}
\label{tab:sim10k}
\end{table}

\noindent
\textbf{Results on Watercolor Dataset~\cite{clipart_watercolor}.} 
The adaptation results on the Watercolor dataset are summarized in Table~\ref{tab:watercolor}. In Watercolor, most of the images contains only one or two objects with less variations of shape and appearance compared with the Clipart dataset.
As reported in SWDA~\cite{SWDA}, the source only model can achieve quite good results with a mAP of 44.6\% and DA-Faster~\cite{adaptiveFasterRCNN} can improve it slightly by only 1.4 points. SWDA \cite{SWDA} performs much better than DA-Faster \cite{adaptiveFasterRCNN}  and obtain a high mAP of 53.5\%. The gain from adaptation is 8.7 points. 
The mAP of our proposed FBC is 53.6\%, which is 0.3\% higher than that of SWDA.
Even without the local feature alignment via adversarial training, our proposed forward-backward cyclic adaptation method can achieve state-of-the-art performance.

\noindent
\textbf{Feature Visualization.}
To visualize the adaptability of our method, we use the Grad-cam~\cite{grad-cam} to show the evidence (heatmap) for the last fully connected layer in the object detectors.
The high value in the heatmap indicates the evidence that why the classifiers make the classification.
Figure~\ref{fig:detection examples} shows the differences of classification evidence before and after adaptation.
As we can see, the adapted detector is able to classify the objects (\eg persons) based on more semantics (\eg faces, necks, joints).
It demonstrates that the adapted detector has addressed the discrepancy on the appearance of real and cartoon objects.
More samples can be found in the supplementary materials.


\subsection{Adaptation from Synthetic to Real Images}
\label{subsec: results sim10k}


As the adaptation from the synthetic images to the real images can potentially reduce the efforts of collecting the real data and labels, we evaluate the adaptation performance in the scenario of Sim10k~\cite{sim10k} to Cityscapes~\cite{cityscapes}.

\noindent
\textbf{Datasets.}
The source domain, Sim10k~\cite{sim10k}, contains synthetic images which are rendered by the computer game Grand Theft Auto (GTA). 
It provides 58,701 bounding box annotations for cars in 10K images.
The target domain, Cityscapes~\cite{cityscapes}, consists of real images captured by car-mounted video camera for driving scenarios.
It comprises 2,975 images for training and 500 images for validation.
We use its training set for adaptation without labels, and validation set for evaluation.
The adaptation is only evaluated on class \textit{car} as Sim10k only provides annotations for car.

\noindent
\textbf{Results.}
Results are shown in Table~\ref{tab:sim10k}.
The reported mAP gain of DA-Faster~\cite{adaptiveFasterRCNN} in its original report (7.8 points) is significantly different from its reproduced gain (-0.4 points) in SWDA~\cite{SWDA}. It implies that a lot of efforts are needed to reproduce the reported results of DA-Faster \cite{adaptiveFasterRCNN}. 
Zhu~\etal\cite{adaptingObjSelective} achieves the best performance with a mAP of 43\%.
Our proposed FBC have a competitive result of mAP, 42.7\%, which is 0.4\% higher than that of SWDA.
But compared with the method by Zhu~\etal\cite{adaptingObjSelective}, our proposed method has a much simpler network architecture and training scheme.

\subsection{Adaptation between Similar Domains}
\label{subsec: resutls foggy}

\begin{table}[t]
\scriptsize
\begin{center}
\begin{tabular}{l|*{9}{p{0.25CM}}}
\hline
Method  &person &rider &car &truck &bus &train &motor &bcycle &mAP \\
\hline
Source Only~\cite{adaptiveFasterRCNN} &17.8 &23.6 &27.1 &11.9 &23.8 &9.1 &14.4 &22.8 &18.8\\
DA-Faster~\cite{adaptiveFasterRCNN}  &25.0 &31.0 &40.5 &22.1 &35.3 &20.2 &20.0 &27.1 &27.6 \\
\hline
Source Only~\cite{adaptingObjSelective} &9.7 &32.2 &44.6 &16.2 &27.0 &9.1 &20.7 &29.7 &26.2\\
Zhu~\etal\cite{adaptingObjSelective}  &33.5 &38.0 &48.5 &26.5 &39.0 &23.3 &28 &33.6 &33.8 \\
\hline
Source Only~\cite{SWDA} &24.1 &33.1 &34.3 &4.1 &22.3 &3.0 &15.3 &26.5 &20.3\\
SWDA~\cite{SWDA}  &29.9 &42.3 &43.5 &24.5 &36.2 &32.6 &30.0 &35.3 &34.3 \\
\hline
Source Only (ours) &22.4 &34.2 &27.2 &12.1 &28.4 &9.5 &20.0 &27.1 &22.9\\
FBC (ours) &31.5 &46.0 &44.3 &25.9 &40.6 &39.7 &29.0 &36.4 &\textbf{36.7} \\
FBC w/o local (ours) &29.0 &37.0 &35.6 &18.9 &32.1 &10.7 &25.0 &31.3 &27.5 \\

\hline
\end{tabular}
\end{center}
\vspace{-5pt}
\caption{Results ($\%$) on the adaptation from Cityscapes~\cite{cityscapes} to FoggyCityscapes Dataset~\cite{foggyCityscapes}.}
\label{tab:foggy}
\vspace{-10pt}
\end{table}

\noindent
\textbf{Datasets.}
The target dataset, FoggyCityscapes~\cite{foggyCityscapes}, is a synthetic foggy dataset where images are rendered from the Cityscapes~\cite{cityscapes}.
The annotations and data splits are the same as the Cityscapes.
The adaptation performance is evaluated on the validation set of FoggyCityscapes.

\noindent
\textbf{Results.}
It can be seen in Table~\ref{tab:foggy} that both SWDA and Zhu~\etal obtain better adaptation results than DA-Faster with mAP of 34.3\% and 33.8\% respectively.
However, compared with the adaptation gain of SWDA (14\%), the gain achieved by Zhu~\etal is only 7.6\%.
Our proposed FBC method outperforms the baseline methods, which boosts the mAP to 36.7\%.
If without the local feature alignment, our proposed method can only obtain limited gain. It is because in this scenario, the main difference between two domains is the local texture.



\noindent
\textbf{t-SNE Visualization.}
We visualize the differences of features before and after adaptation via t-SNE visualization~\cite{tSNE} in Figure~\ref{fig: tsne}.
The features are output from the ROI pooling layer and 100 images are randomly selected.
After adaptation, the distributions of source and target features are well aligned with regard to the object classes.
More importantly, as shown in Figure~\ref{fig: tsne}(b), different classes are better distinguished and more target objects are detected for each class after adaptation.
This demonstrates the effectiveness of our proposed adaptation method for object detection.

\begin{figure}[t]
\begin{center}
   \includegraphics[width=0.8\linewidth]{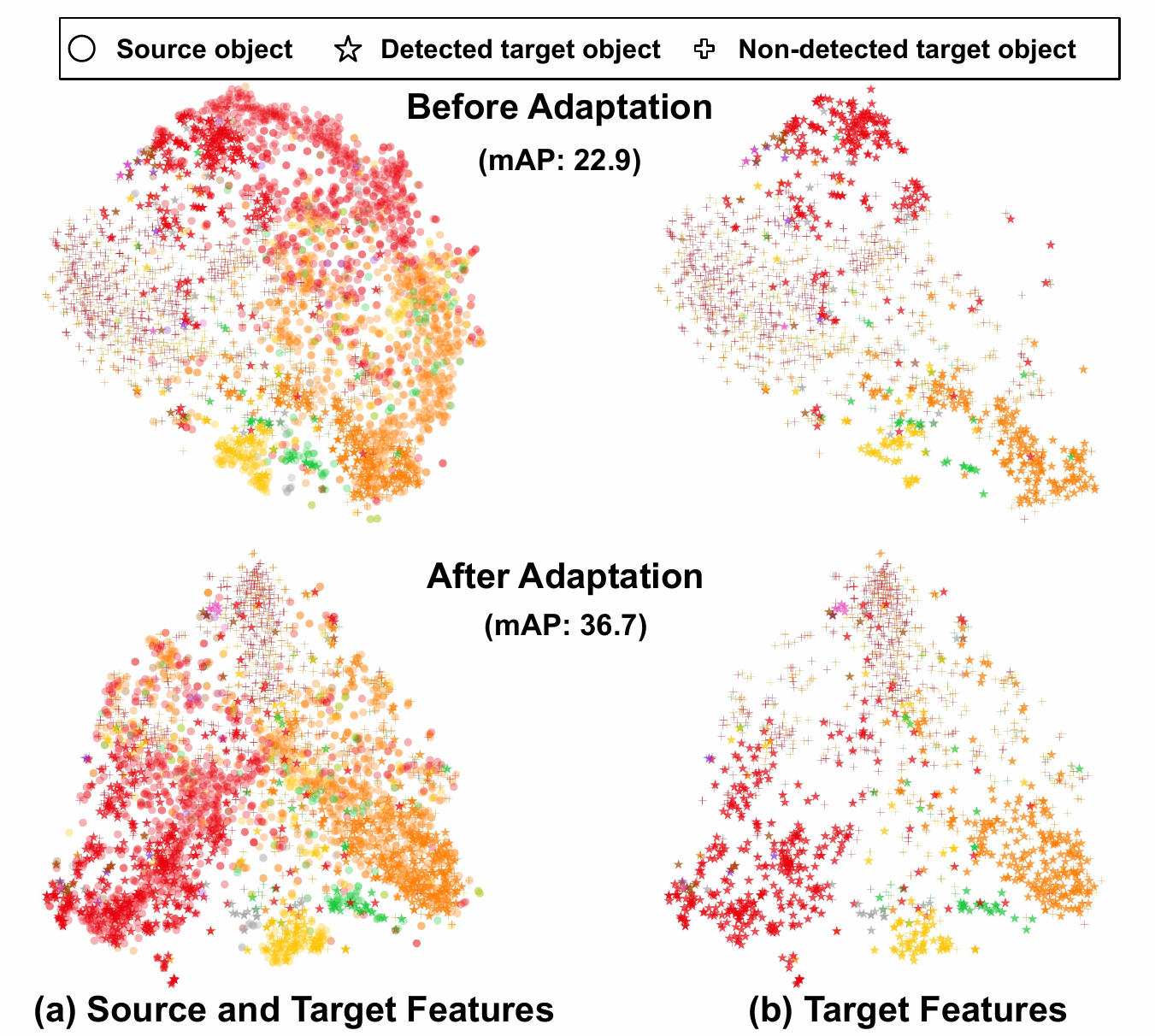}
\end{center}
\vspace{-10pt}
   \caption{t-SNE visualization of features before and after domain adaptation from Cityscape to FoggyCityScape. Different colors represent different classes. Target features are displayed alone on the right for better visualization.}
   \label{fig: tsne}
   \vspace{-10pt}
\end{figure}
\section{Conclusions}

We addressed the unsupervised domain adaptation for object detection task where the target domain does not have labels. A forward-backward cyclic adaptation method is proposed. This method was based on the intuition that domain invariance of category level semantics could be learnt when the gradient directions of source and target were aligned. Theoretical analysis was presented to show that the proposed method achieved the gradient alignment goal. Local feature alignment via adversarial training was performed for learning domain-invariance of holistic color/textures. Furthermore, we proposed a domain diversity constraint to penalize confident source-specific learning and intrigue target-specific learning via entropy regularization.



\onecolumn
%
%
%
\renewcommand{\thesection}{\Alph{section}}
%
%
%
%

In the following supplementary material, 
we fist provide detailed theoretical analysis to illustrate how our proposed Forward and Backward Cyclic Adaptation (FBC) in Algorithm 1 approximates the objective function of gradient alignment (to Eq.3 in our main submission). 
Details are shown in Section~\ref{Deriving the Objective Function}.
We then provide ablation studies in Section~\ref{Ablation Studies}.
Section~\ref{Feature Visualization} demonstrates more examples of feature visualization on the Watercolor~\cite{clipart_watercolor}.

\section{Deriving the Objective Function}
\label{Deriving the Objective Function}

We detail the theoretical analysis in the main submission to show how the proposed algorithm approximates the objective function of gradient alignment.
We follow the conventions in Reptile~\cite{reptile} and demonstrate the gradient computations during the training.
In Reptile~\cite{reptile}, they effectively extrapolated the gradient with a number of steps taken.
Let us first denote the terms following~\cite{reptile,MaxTransferMinInterference}:
\begin{align}
g_i &= \frac{\partial \mathcal{L}_i(\theta_i)}{\partial \theta_i} \quad (\textrm{gradient obtained during SGD}) \textrm{,}\\
\theta_{i+1} &= \theta_i - \alpha g_i \quad (\textrm{squence of parameter vectors)} \textrm{,}\\
\bar{g}_i &= \frac{\partial \mathcal{L}_i(\theta_i)}{\partial \theta_1} \quad (\textrm{gradient at initial point}) \textrm{,}\\
g^j_i &= \frac{\partial \mathcal{L}_i(\theta_i)}{\partial \theta_j} \quad (\textrm{gradient evaluated at point i with respect to parameters j}) \textrm{,}\\
\bar{H}_i &= \frac{\partial^2 \mathcal{L}_i(\theta_i)}{\partial \theta^2_1} \quad (\textrm{Hessian at initial point})\textrm{,} \\
H^j_i &= \frac{\partial^2 \mathcal{L}_i(\theta_i)}{\partial \theta^2_j} \quad (\textrm{Hessian evaluated at point i with respect to parameters j}) \textrm{,}
\end{align} 
where the $\alpha$ is learning rate and $\mathcal{L}_i$ is the loss function on the samples for each gradient updates.

According to the Taylor's theorem, we have the SGD gradients as follows:
\begin{align}
g_i = L^\prime_i(\theta_1) &= \mathcal{L}^\prime_i(\theta_1) + \mathcal{L}^{\prime\prime}_i(\theta_i-\theta_1) + O(\lVert \theta_i - \theta_1 \rVert ^2) \textrm{,} \\
& = \bar{g}_i + \bar{H}_i(\theta_i-\theta_1) + O(\lVert \theta_i - \theta_1 \rVert ^2) \quad (\textrm{using definition of} \: \bar{g}_i, \bar{H}_i ) \textrm{,}\\
& = \bar{g}_i - \alpha \bar{H}_i \sum^{i-1}_{j=1} g_j + O(\lVert \theta_i - \theta_1 \rVert ^2) \quad (\textrm{using gradient updates} \: \theta_i - \theta_1 = -\alpha \sum^{i-1}_j g_j ) \textrm{,}\\
& = \bar{g}_i - \alpha \bar{H}_i \sum^{i-1}_{j=1} \bar{g}_j + O(\lVert \theta_i - \theta_1 \rVert ^2) \quad (\textrm{using} \: g_j = \bar{g}_j + O(\lVert \theta_i - \theta_1 \rVert ^2) ) \textrm{.}
\label{eq:taylor}
\end{align}
If we consider there are two steps of parameter updates with stochastic gradient descent (SGD), where the gradient of the first step is $g_1$ and the one of second step is $g_2$. According to the Eq.~\ref{eq:taylor}, we have
\begin{align}
g_1 &= \bar{g}_1 \textrm{,} \\
g_2 &= \bar{g}_2 - \alpha \bar{H}_2 \bar{g}_1 +O(\lVert \theta_i - \theta_1 \rVert ^2) \textrm{.}
\end{align}
Then, the overall gradient of the two SGD steps is 
\begin{align}
g = g_1 + g_2 = \bar{g}_1 + \bar{g}_2 - \alpha \bar{H}_2 \bar{g}_1 + O(\lVert \theta_i - \theta_1 \rVert ^2) \textrm{.}
\end{align}
In Reptile~\cite{reptile}, they noted that
\begin{align}
\epsilon[\bar{H}_2\bar{g}_1] = \epsilon[\bar{H}_1\bar{g}_2] = \frac{1}{2} \epsilon [\bar{H}_2\bar{g}_1 + \bar{H}_1\bar{g}_2] = \frac{1}{2} \epsilon [\frac{\partial}{\partial \theta_1} (\bar{g}_1 \bar{g}_2)]\textrm{,}
\end{align}
where the $\epsilon$ is the expected loss. Therefore, the overall expected loss is 
\begin{align}
\epsilon[g] = \epsilon[\bar{g}_1] + \epsilon[\bar{g}_2] -  \frac{1}{2} \alpha \epsilon [\frac{\partial}{\partial \theta_1} (\bar{g}_1 \bar{g}_2)]\textrm{.}
\label{eq:expected error g}
\end{align}

In our work, we aim to address the domain adaptation problem for object detection.
In our proposed forward and backward cyclic adaptation (Algorithm 1), we train the network with episodic training. 
In each episode, similar to the two-step SGD updates discussed above, we first perform the backward hopping on labeled source samples $\{\mathcal{X}_S, \mathcal{Y}_S\}$ to obtain the parameters $\theta_S$, and then we initialize the forward passing with $\theta_S$ and train the network with pseudo labeled target samples $\{\mathcal{X}_T, \mathcal{\hat{Y}}_T\}$, obtaining the updated parameters $\theta_T$.
The shared model $\theta$ is updated by $\theta_S$ and $\theta_T$ sequentially.
We can consider the gradient of forward passing, $g_S$, as $g_1$, and similarly $g_T$ as $g_2$. 
Then we can substitute $g_S$ and $g_T$ to Eq.~\ref{eq:expected error g}:
\begin{align}
\mathbb{E}[g_{e}] = \mathbb{E}[\bar{g}_S] + \mathbb{E}[\bar{g}_T] -  \frac{1}{2} \alpha \epsilon [\frac{\partial}{\partial \theta_S} (\bar{g}_S\bar{g}_T)]\textrm{ ,}
\end{align}
where $\mathbb{E}$ is the expected loss.
The above equation shows that the training of our proposed adaptation method (Algorithm 1) is approximating the objective of gradient alignment:
\begin{align}
\min_{\theta_S, \theta_T} \mathcal{L}_{\theta_S}(\mathcal{X}_S, \mathcal{Y}_S)+\mathcal{L}_{\theta_T}(\mathcal{X}_T, \mathcal{\hat{Y}}_T) - \alpha \frac{\partial \mathcal{L}_{\theta_S}(\mathcal{X}_S, \mathcal{Y}_S)}{\partial \theta_S} \cdot \frac{\partial \mathcal{L}_{\theta_T}(\mathcal{X}_T, \mathcal{\hat{Y}}_T)}{\partial \theta_T} \textrm{ .}
\end{align}


%


\section{Ablation Studies}
\label{Ablation Studies}
In this section, we evaluate the effects of the different components in our proposed adaptation method. As shown in the Eq.11 and Eq.12 in our submission, our overall objective function is 
\begin{align*}
\min_{\theta} \mathcal{L} &= \mathcal{L}_{inv}(\mathcal{X}_S, \mathcal{Y}_S, \mathcal{X}_T) + \gamma \mathcal{L}_{div}(\mathcal{X}_S, \mathcal{X}_T) \\
&= \mathcal{L}_g(\mathcal{X}_S, \mathcal{Y}_S, \mathcal{X}_T) + \lambda\mathcal{L}_{adv}(\mathcal{X}_S,  \mathcal{X}_T) + \gamma \mathcal{L}_{div}(\mathcal{X}_S, \mathcal{X}_T)\textrm{ ,}
\end{align*}
where $\mathcal{L}_{g}$ is the loss of gradient alignment,  $\mathcal{L}_{adv}$ is the loss of local feature alignment via adversarial training and $\mathcal{L}_{div}$ is the loss of domain-diversity.
$\lambda$ and $\gamma$ are the hyperparameters and we set $\lambda=0.5$ and $\gamma=0.1$ for all the experiments in this work. 

In the following sections, we use G, L, and D to indicate gradient alignment, local feature alignment and domain diversity, respectively.

\subsection{Effects of Gradient Alignment}
To evaluate the effects of gradient alignment, we perform the forward-backward cyclic method (FBC) on four different cross-domain scenarios with gradient alignment only.
The results are shown in Table~\ref{tab:clipart} -~\ref{tab:foggy}.
In the adaptation scenarios, PASCAL~\cite{PASCAL}-to-Clipart~\cite{clipart_watercolor} (in Table~\ref{tab:clipart}) and PASCAL-to-Watercolor~\cite{clipart_watercolor} (in Table~\ref{tab:watercolor}), the FBC with gradient alignment can achieve better adaptation results than the FBC with local feature alignment only.
It is because the domain discrepancy in these two adaptation scenarios is large, \ie adapting real objects to cartoon or watercolor objects.
This indicates that gradient alignment has its superiority in aligning high-level semantics.


However, in the adaptation scenarios,  Sim10k~\cite{sim10k}-to-Cityscapes~\cite{cityscapes}  (in Table~\ref{tab:sim10k} and Cityscapes~\cite{cityscapes}-to-FoggyCityscapes~\cite{foggyCityscapes} (in Table~\ref{tab:foggy} , the domain discrepancy between two domain are mainly in the low-level features, \eg textures and colors.
Therefore, in these scenarios, the FBC with gradient alignment only can achieve limited gain on mAP, compared witht he FBC with local feature alignment.
It is more evident in Cityscapes-to-FoggyCityscapes, where the foggy images are rendered from the real images.
However, the FBC with gradient alignment only is still $4.6\%$ higher than the source only model (in Table~\ref{tab:foggy}).
Although the FBC with local feature alignment can obtain a high mAP with $33.7\%$, in combination with gradient alignment and domain diversity, the mAP can be boosted to $36.7\%$.

\subsection{Effects of Local Feature Alignment}
The local feature alignment is conducted via adversarial training, which aligns the marginal feature distributions between the source and target domains.
As discussed in the main submission, the alignment of marginal feature distributions does not perform well when the domain discrepancy is large.
This is also demonstrated in our experiments.
In Table~\ref{tab:clipart} and Table~\ref{tab:watercolor}, the FBC with local feature alignment only does not perform better than the gradient alignment, when the domain discrepancy is large.
But when the domain discrepancy is small, \ie in low-level semantics, the FBC with local feature alignment demonstrates its superiority, as shown in Table~\ref{tab:sim10k} and Table~\ref{tab:foggy}.

It is worthy to mention that the gradient alignment and local feature alignment are complementary, as gradient alignment can achieve category-level alignment for high-level semantics and local feature alignment via adversarial training has its advantages for aligning low-level semantics.
The combination of these two alignment and domain diversity can achieve the state-of-the-art performance.

\subsection{Effects of Domain Diversity}

Here we evaluate the effects of the domain-diversity.
As shown in Table~\ref{tab:clipart} - Table~\ref{tab:foggy}, the domain diversity can consistently improve the adaptation results.


\begin{table*}[h]
\scriptsize
\begin{center}
\begin{tabular}{l|c c c|p{0.2cm}p{0.22cm}*{17}{p{0.21cm}}p{0.29cm}|p{0.4cm}}
\hline
Method &G &L &D &aero &bcycle &bird &boat &bottle &bus &car &cat &chair &cow &table &dog &hrs &motor &prsn &plnt &sheep &sofa &train &tv & mAP \\
\hline
Source Only (ours) & & & &24.2 &47.1 &24.9 &17.7 &26.6 &47.3 &30.4 &11.9 &36.8 &26.4 &10.1 &11.8 &25.9 &74.6 &42.1 &24.0 &3.8 &27.2 &37.9 &29.9 &29.5 \\
\hline
\multirow{2}{*}{FBC (ours)} 
&$\checkmark$ & & &28.8 &64 &21.1 &19.1 &39.7 &60.7 &29.5 &14.2 &46.4 &29.3 &21.8 &8.9 &28.8 &72.7 &51.3 &32.9 &12.8 &28.1 &52.7 &49.5 &35.6 \\
&$\checkmark$ & &$\checkmark$ &32.1 &57.6 &24.4 &23.7 &34.1 &59.3 &32.2 &9.1 &40.3 &41.3 &27.8 &11.9 &30.2 &72.9 &48.8 &38.3 &6.1 &33.1 &46.5 &48 &35.9 \\
& &$\checkmark$ & &31.8 &53.0 &21.3 &25.0 &36.1 &55.9 &30.4 &11.6 &39.3 &21.0 &9.4 &14.5 &32.4 &79.0 &44.9 &37.8 &6.2 &35.6 &43.0 &53.5 &34.1 \\
&$\checkmark$ &$\checkmark$ &$\checkmark$ &43.9 &64.4 &28.9 &26.3 &39.4 &58.9 &36.7 &14.8 &46.2 &39.2 &11.0 &11.0 &31.1 &77.1 &48.1 &36.1 &17.8 &35.2 &52.6 &50.5 &\textbf{38.5} \\
\hline
\end{tabular}
\end{center}
\caption{The results ($\%$) on the adaptation from PASCAL~\cite{PASCAL} to Clipart Dataset~\cite{clipart_watercolor}.}
\label{tab:clipart}
\end{table*}

\begin{table}[h]
\scriptsize
\begin{center}
\begin{tabular}{l|c c c|*{6}{p{0.25cm}}|p{0.5cm}}
\hline
Method &G &L &D &bike &bird &car &cat &dog &prsn &mAP \\
\hline
Source Only (ours) & & & &66.7 &43.5 &41 &26.0 &22.9 &58.9 &43.2 \\
\hline
\multirow{2}{*}{FBC (ours)} 
&$\checkmark$ & &   &90.9	&46.5	&51.3	&33.2	&29.5	&65.9	&52.9  \\
&$\checkmark$ & &$\checkmark$ &88.7 &48.2 &46.6 &38.7 &35.6 &64.1 &\textbf{53.6}\\
& &$\checkmark$ &  &89.0	&47.2	&46.1	&39.9	&27.7	&65.0	&52.5 \\
 &$\checkmark$ &$\checkmark$ &$\checkmark$ &90.9 &47.7 &46.0 &38.7 &31.8 & 66.7 &\textbf{53.6}  \\
\hline
\end{tabular}
\end{center}
\caption{The results ($\%$) on the adaptation from PASCAL~\cite{PASCAL} to Watercolor Dataset~\cite{clipart_watercolor}.}
\label{tab:watercolor}
\end{table}

%
%

\begin{table}[h]
\scriptsize
\begin{center}
\begin{tabular}{l|c c c|c}
\hline
Method  &G &L &D &AP on Car  \\
\hline
Source Only (ours) & & & &31.2 \\
\hline
\multirow{2}{*}{FBC (ours)} 
&$\checkmark$ & & &38.2  \\
&$\checkmark$ & &$\checkmark$ &39.2 \\
& &$\checkmark$ &  &41.4 \\
&$\checkmark$ &$\checkmark$ &$\checkmark$ &\textbf{42.7}  \\
\hline
\end{tabular}
\end{center}
\caption{The results ($\%$) on the adaptation from Sim10k~\cite{sim10k} to Cityscapes Dataset~\cite{cityscapes}.}
\label{tab:sim10k}
\end{table}

\begin{table}[h]
\scriptsize
\begin{center}
\begin{tabular}{l|c c c|*{9}{p{0.25CM}}}
\hline
Method  &G &L &D &person &rider &car &truck &bus &train &motor &bcycle &mAP \\
\hline
Source Only (ours) & & & &22.4 &34.2 &27.2 &12.1 &28.4 &9.5 &20.0 &27.1 &22.9\\
\hline
\multirow{2}{*}{FBC (ours)} 
&$\checkmark$ & & &25.8	&35.6	&35.5	&18.4	&29.6	&10.0	&24.5	&30.3	&26.2 \\
&$\checkmark$ & &$\checkmark$&29.0 &37.0 &35.6 &18.9 &32.1 &10.7 &25.0 &31.3 &27.5 \\
& &$\checkmark$ &  &31.6	&45.1	&42.6	&26.4	&37.8	&22.1	&29.4	&34.6	&33.7\\
&$\checkmark$ &$\checkmark$ &$\checkmark$ &31.5 &46.0 &44.3 &25.9 &40.6 &39.7 &29.0 &36.4 &\textbf{36.7} \\
\hline
\end{tabular}
\end{center}
\caption{Results ($\%$) on the adaptation from Cityscapes~\cite{cityscapes} to FoggyCityscapes Dataset~\cite{foggyCityscapes}.}
\label{tab:foggy}
\end{table}
\section{More Implementation Details}
In this section, we provide more implementation details of our experiments.

\noindent
\textbf{Details of Local Feature Alignment. }
In this work, we utilize the Gradient Reversal Layer (GRL) proposed by Ganin and Lempitsky~\cite{GRL} for adversarial training.
Following SWDA~\cite{SWDA}, we extract local features from low-level layer as input of the domain classifier $D$
and the least-squares loss~\cite{LSGAN,cyclegan}.
The domain classifier is the same as the local domain classifier in SWDA, which consists of three layered convolutional layers with kernel size as 1.

For the local features, the features output from \textit{conv3-3} are extracted in the case of VGG16 model and the features output from the last \textit{res3c} layer are extracted in ResNet101 model.
The name of the layer follows the prototxt in Caffe~\cite{caffe}.

\section{Feature Visualization}
\label{Feature Visualization}

\begin{figure}[t]
\begin{center}
   \includegraphics[width=0.9\linewidth]{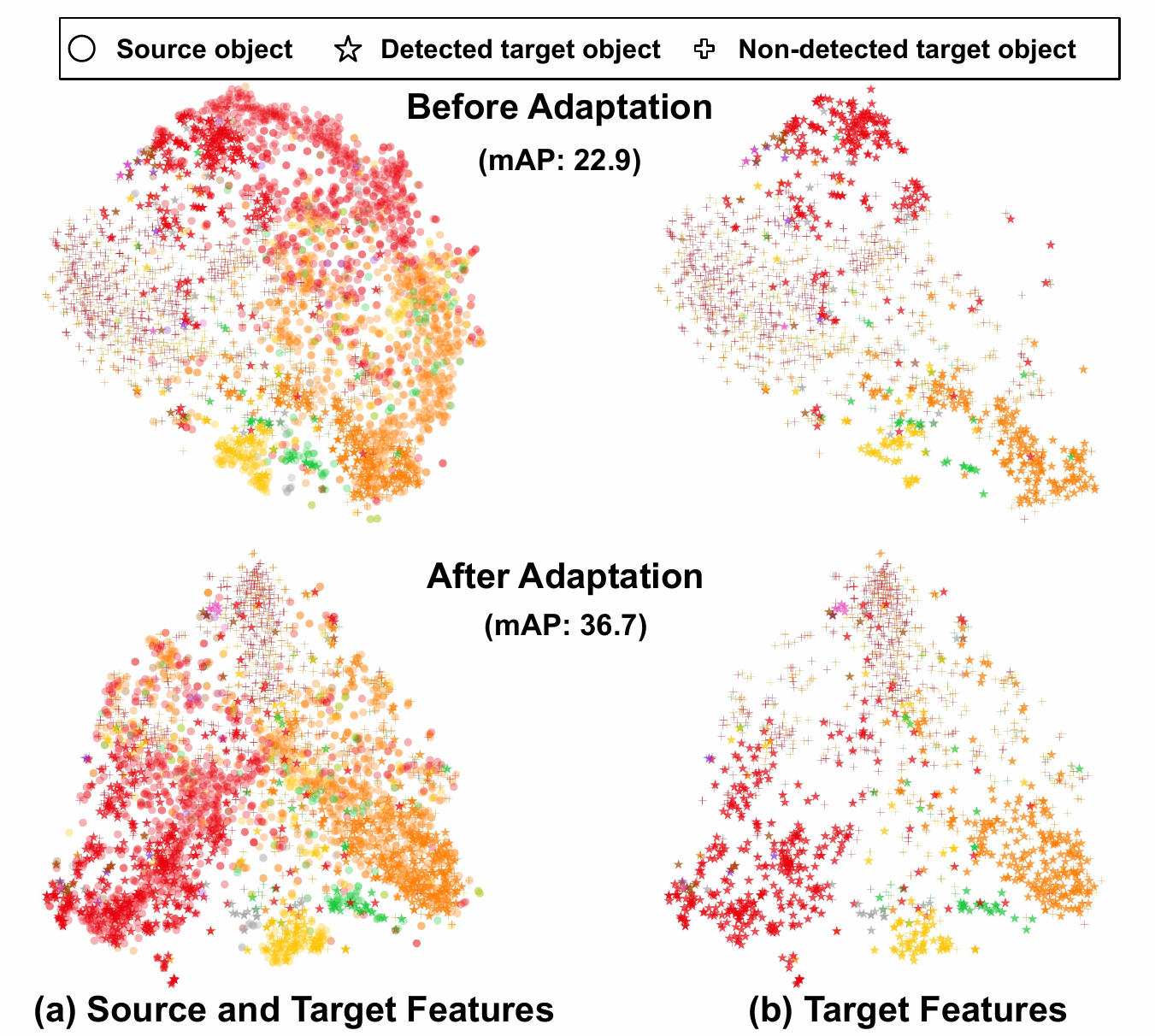}
\end{center}
   \caption{t-SNE visualization of features before and after domain adaptation from Cityscape to FoggyCityScape. Different colors represent different classes. Target features are displayed alone on the right for better visualization.}
   \label{fig: tsne}
\end{figure}

\begin{figure}[t]
\centering
\includegraphics[width=0.7\linewidth]{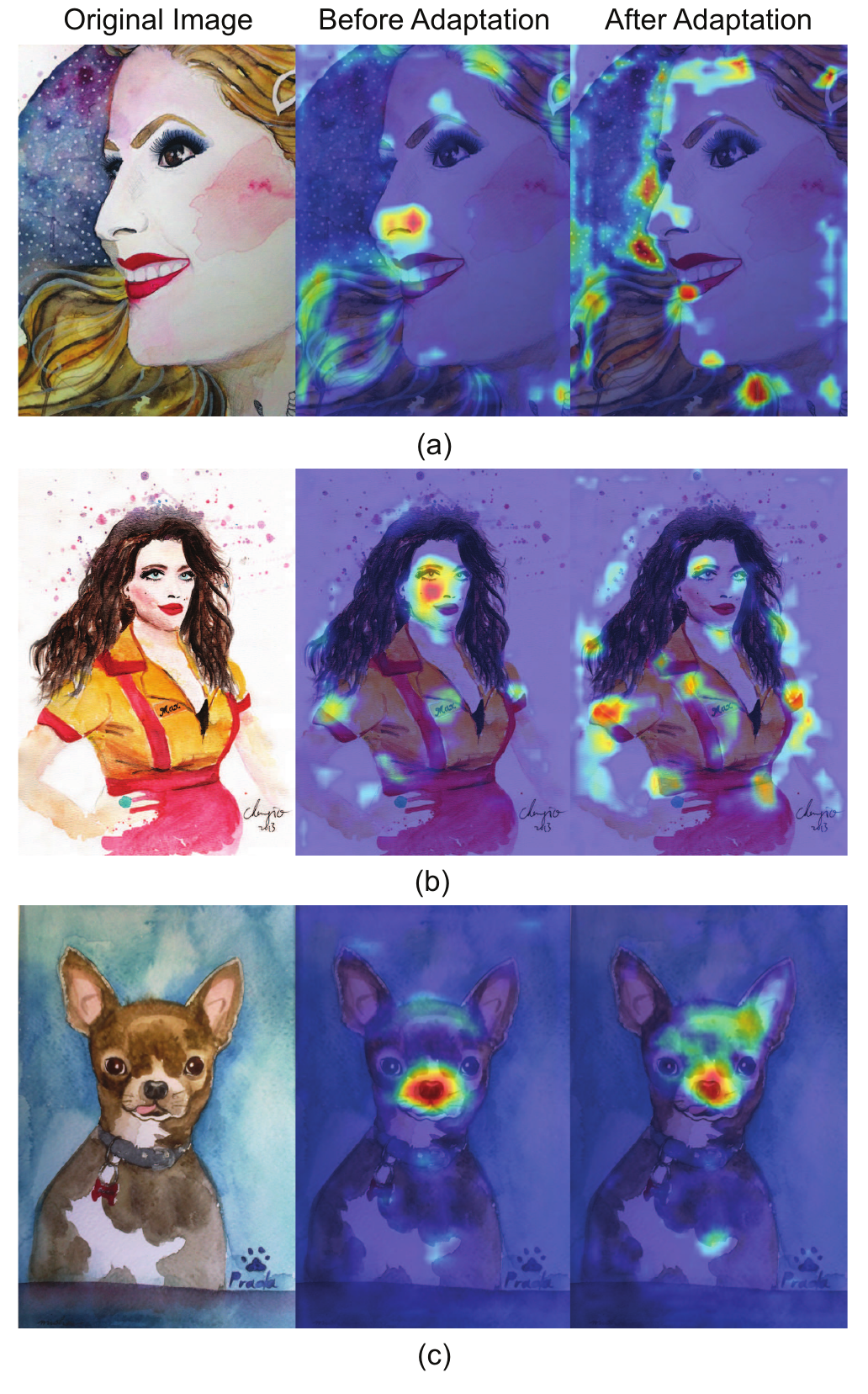}
\caption{Feature visualization shows the evidence for improvements in classifiers before and after domain adaptation using Grad-cam~\cite{grad-cam} on the Watercolor dataset~\cite{clipart_watercolor}. The images in the middle column show the attention for the classifier before adaptation and the one on the right show the attention for the classifier after adaptation. This figure demonstrates that the adapted detector utilizes more semantics to classify the objects, which indicates the effectiveness of our proposed domain adaptation method. }
\label{fig:grad cam watercolor}
\end{figure}

%

\twocolumn
\newpage
{\small
\bibliographystyle{ieee_fullname}
\bibliography{reference}
}

\end{document}